\documentclass[times,twocolumn,final]{elsarticle}

\usepackage{prletters}
\usepackage{framed,multirow}
\usepackage{amssymb}
\usepackage{latexsym}
\usepackage{url}
\usepackage{xcolor}
\usepackage{times}
\usepackage{epsfig}
\usepackage{graphicx}
\usepackage{amsmath}
\usepackage{gensymb}
\usepackage{caption}
\usepackage{algorithm,algorithmic}%
\usepackage{color,colortbl}
\usepackage{rotating}
\usepackage{tikz}
\usepackage{pgfplots}
\usepackage{xspace}
\usepackage{hyperref}

\usepackage[printwatermark]{xwatermark}
\newwatermark[allpages,color=gray!30,angle=45,scale=3,xpos=0,ypos=0]{PRE-PRINT}

\definecolor{newcolor}{rgb}{.8,.349,.1}

\journal{Pattern Recognition Letters}

\DeclareMathOperator*{\argmin}{arg\,min}

\def\eg{\emph{e.g.}} 
\def\ie{\emph{i.e.}} 
\def\etal{\emph{et al.}~}

\begin{document}

\begin{frontmatter}

\title{Mesh-based Camera Pairs Selection\\ and Occlusion-Aware Masking for Mesh Refinement}

\author[1]{Andrea \snm{Romanoni}\corref{cor1}} 
\cortext[cor1]{Corresponding author:}
\ead{andrea.romanoni@polimi.it}
\author[1]{Matteo \snm{Matteucci} }

\address[1]{Politecnico di Milano, Italy}

\received{}
\finalform{}
\accepted{}
\availableonline{}
\communicated{}

\begin{abstract}
Many Multi-View-Stereo algorithms extract a 3D mesh model of a scene, after fusing depth maps into a volumetric representation of the space. Due to the limited scalability of such representations, the estimated model does not capture fine details of the scene. Therefore a mesh refinement algorithm is usually applied; it improves the mesh resolution and accuracy by minimizing the photometric error induced by the 3D model into pairs of cameras. The choice of these pairs significantly affects the quality of the refinement and usually relies on sparse 3D points belonging to the surface. Instead, in this paper, to increase the quality of pairs selection, we exploit the 3D model (before the refinement) to compute five metrics: scene coverage, mutual image overlap, image resolution, camera parallax, and a new symmetry term. To improve the refinement robustness, we also propose an explicit method to manage occlusions, which may negatively affect the computation of the photometric error. The proposed method takes into account the depth of the model while computing the similarity measure and its gradient. We quantitatively and qualitatively validated our approach on publicly available datasets against state of the art reconstruction methods.
\end{abstract}

\begin{keyword}
\MSC 41A05\sep 41A10\sep 65D05\sep 65D17
\KWD Keyword1\sep Keyword2\sep Keyword3

%% MSC codes here, in the form: \MSC code \sep code
%% or \MSC[2008] code \sep code (2000 is the default)
\end{keyword}

\makeatletter
\let\@oldmaketitle\@maketitle% Store \@maketitle
\renewcommand{\@maketitle}{\@oldmaketitle
 \vspace{-1.5em}
 
\centering
\setlength{\tabcolsep}{1px}
    \begin{tabular}{ccc}
    \includegraphics[width=0.3\textwidth]{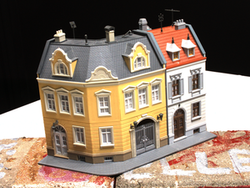}&
    \includegraphics[width=0.3\textwidth]{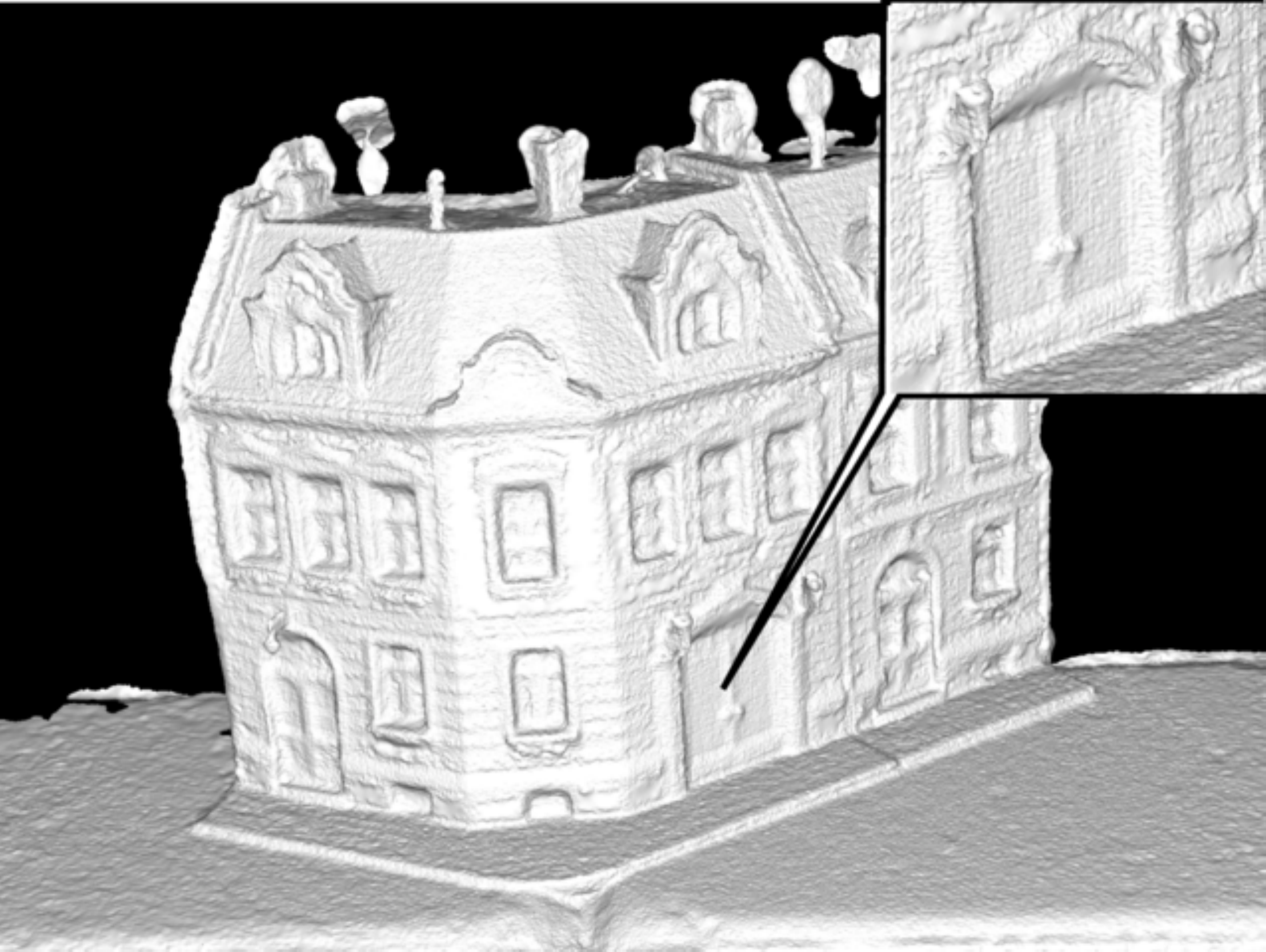}&
    \includegraphics[width=0.3\textwidth]{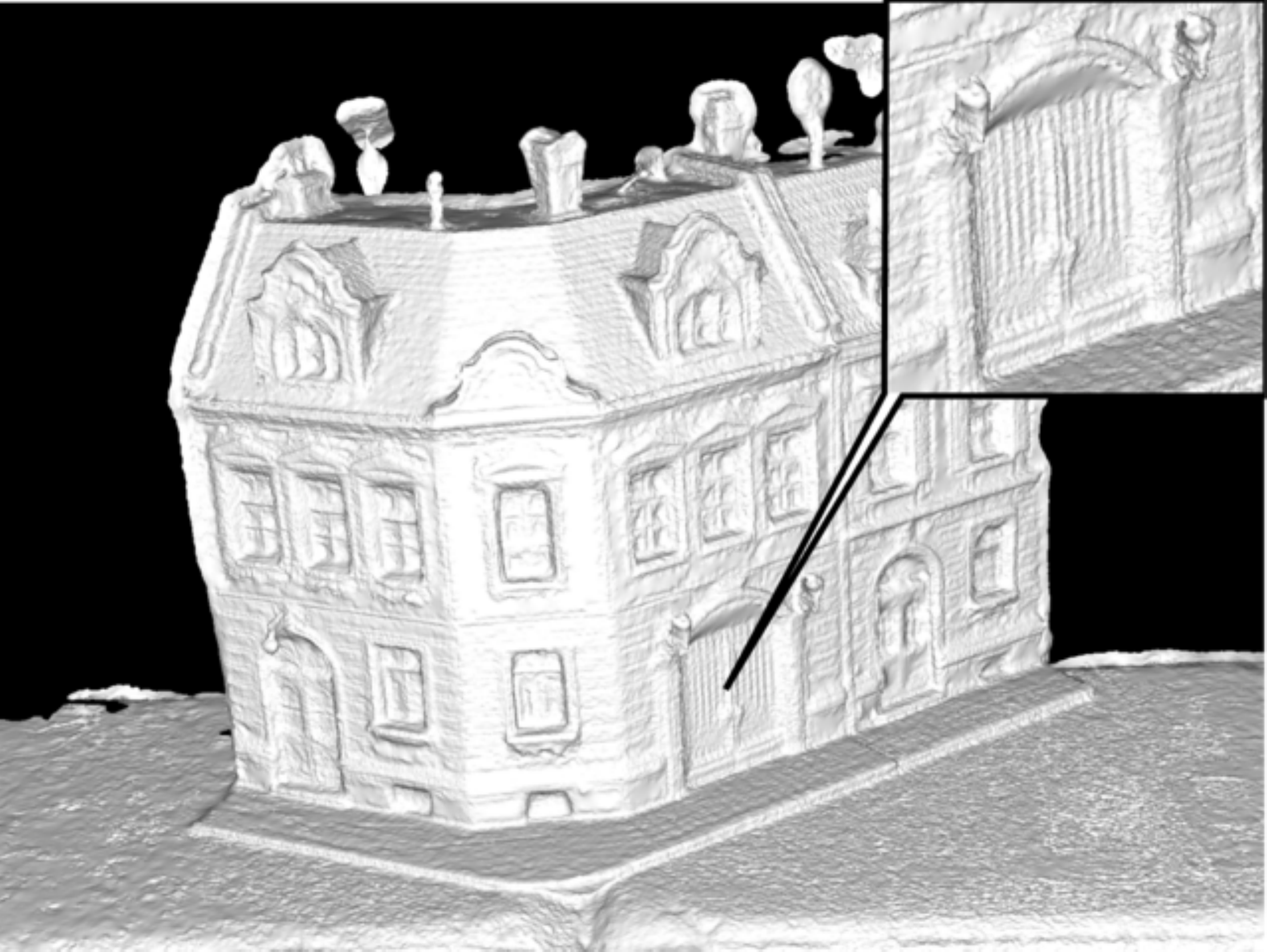}\\
    (a)&(b)&(c)\\
    \end{tabular}
    \captionof{figure}{Mesh Refinement: (a) original image, (b) initial mesh (c) refined mesh.}
\label{fig:exw}
%  \captionof{figure}{While solid paints (left top).}\label{fig:fig1}
    \bigskip
    }
\makeatother

\definecolor{Gray}{gray}{0.9}
\end{frontmatter}

\section{Introduction}
Multi-View Stereo (MVS) algorithms recover the 3D model of a scene captured by a set of images.
Boosted by the benchmarks proposed in \cite{strecha2008,seitz_et_al06,aanaes2016large} and the enhancements in hardware capabilities, several works proposed several accurate and efficient MVS methods.%, which often output a 2D mesh embedded in the 3D space.

A well established MVS pipeline, first proposed by {Vu \etal \cite{vu_et_al_2012}}, estimates the camera positions with Structure-from Motion (SfM) \cite{openMVG,wu2011multicore}; then it applies plane sweeping \cite{collins1996space} or depth map fusion \cite{hiep2009towards} to recover a dense point cloud representation of the scene. 
This pipeline builds a volumetric partitioning of the space in which the camera to point visibility rays are exploited to estimate free and occupied space (or an implicit representation of the model, such as Truncated Signed Distance Function); the free-occupied space boundary (or the zero crossing surface) is the model of the observed scene, usually represented by means of a mesh.
To obtain a very accurate reconstruction, the last step of the pipeline is a refinement algorithm; it minimizes the photometric error induced by such mesh in pairs of cameras.

A fundamental aspect of a mesh refinement algorithm, and in general a Multi-View Stereo method, is the choice of the camera pairs used to compute and minimize the photometric error.
It is well known that too narrow cameras imply noisy reconstruction results. On the other hand, images captured by cameras too far from each other could have limited overlap, \ie, the region of the scene perceived by both cameras is small.
The right choice of these pairs leads to a coherent computation of the photometric error and, as a consequence, an effective gradient descent minimization.
The most widespread Multi-View Stereo methods select for each camera a pairing view among the others by relying on several factors. Li \etal \cite{li2010bundled} and Ebner \etal \cite{ebner2017fully} evaluate the baseline and the angle of the principal viewing direction between the cameras; instead, other methods \cite{goesele2007multi,tola2012efficient,shen2012depth,lou2014accurate}  consider the SfM 3D points and take into account the average angle between the camera-to-point viewing rays, the baseline among views and the scale.
Vogiatzis \etal \cite{vogiatzis2005multi} leverage on similar metrics to filter out unreliable photometric measures, adopted to estimate the 3D model.

While pixel-wise camera pair selection has been addressed to estimate accurate point clouds in \cite{zheng2014patchmatch,schonberger2016pixelwise,romanoni2019tapa}, mesh refinement literature has always limited the choice to the pairs of cameras sharing the visibility of the highest number of 3D points with sufficient parallax.
% Moreover, they does not take into account that, depending on the camera pair configuration with respect to the surface, the image error could have different influence on the refinement, \eg ,higher an image error in a camera that perceives a slanted surface would affect the mesh evolution more significantly w.r.t. an error in a front-facing camera.

\begin{figure}[t]
\centering
  \def\svgwidth{0.7\columnwidth}
  %% Creator: Inkscape inkscape 0.92.3, www.inkscape.org
%% PDF/EPS/PS + LaTeX output extension by Johan Engelen, 2010
%% Accompanies image file 'occl.pdf' (pdf, eps, ps)
%%
%% To include the image in your LaTeX document, write
%%   \input{<filename>.pdf_tex}
%%  instead of
%%   \includegraphics{<filename>.pdf}
%% To scale the image, write
%%   \def\svgwidth{<desired width>}
%%   \input{<filename>.pdf_tex}
%%  instead of
%%   \includegraphics[width=<desired width>]{<filename>.pdf}
%%
%% Images with a different path to the parent latex file can
%% be accessed with the `import' package (which may need to be
%% installed) using
%%   \usepackage{import}
%% in the preamble, and then including the image with
%%   \import{<path to file>}{<filename>.pdf_tex}
%% Alternatively, one can specify
%%   \graphicspath{{<path to file>/}}
%% 
%% For more information, please see info/svg-inkscape on CTAN:
%%   http://tug.ctan.org/tex-archive/info/svg-inkscape
%%
\begingroup%
  \makeatletter%
  \providecommand\color[2][]{%
    \errmessage{(Inkscape) Color is used for the text in Inkscape, but the package 'color.sty' is not loaded}%
    \renewcommand\color[2][]{}%
  }%
  \providecommand\transparent[1]{%
    \errmessage{(Inkscape) Transparency is used (non-zero) for the text in Inkscape, but the package 'transparent.sty' is not loaded}%
    \renewcommand\transparent[1]{}%
  }%
  \providecommand\rotatebox[2]{#2}%
  \newcommand*\fsize{\dimexpr\f@size pt\relax}%
  \newcommand*\lineheight[1]{\fontsize{\fsize}{#1\fsize}\selectfont}%
  \ifx\svgwidth\undefined%
    \setlength{\unitlength}{188.53064346bp}%
    \ifx\svgscale\undefined%
      \relax%
    \else%
      \setlength{\unitlength}{\unitlength * \real{\svgscale}}%
    \fi%
  \else%
    \setlength{\unitlength}{\svgwidth}%
  \fi%
  \global\let\svgwidth\undefined%
  \global\let\svgscale\undefined%
  \makeatother%
  \begin{picture}(1,0.52751341)%
    \lineheight{1}%
    \setlength\tabcolsep{0pt}%
    \put(0,0){\includegraphics[width=\unitlength,page=1]{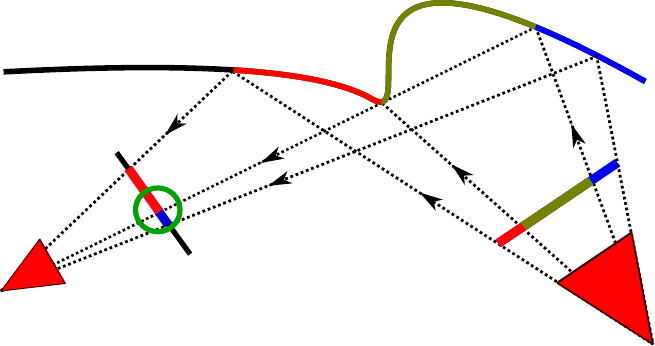}}%
    \put(0.2890134,0.10159441){\color[rgb]{0,0,0}\makebox(0,0)[lt]{\lineheight{0}\smash{\begin{tabular}[t]{l}$I$\end{tabular}}}}%
    \put(0.65338573,0.14156662){\color[rgb]{0,0,0}\makebox(0,0)[lt]{\lineheight{0}\smash{\begin{tabular}[t]{l}$J$\end{tabular}}}}%
    \put(0.92711523,0.46865179){\color[rgb]{0,0,0}\makebox(0,0)[t]{\lineheight{0}\smash{\begin{tabular}[t]{c}$S$\end{tabular}}}}%
  \end{picture}%
\endgroup%

\caption{Effect of the occlusion in the error computation}
\label{fig:occl}
\end{figure}

A second issue which affects mesh refinement and we address in this paper is related to model occlusions. 
For instance, in Fig. \ref{fig:occl} image $J$ projects in $I$ through $S$; the patch in the green circle contains a discontinuity. In this case, while computing the projection error in the green patch corresponding to a pixel in the red region, state-of-the-art methods consider both the information from the blue and red (non-adjacent) parts of $S$. 
This issue has been considered only when dealing with generative methods \cite{delaunoy_et_al_08} or with a simple heuristic avoiding to evolve the mesh just in correspondence of edge segments joining visible and non-visible facets (\cite{li2016efficient}).

For these reasons, in this paper we propose three contributions:
\begin{itemize}
\item A pairwise camera selection method exploiting the knowledge of an approximate model of the scene. It minimizes an energy function defined over the surface instead of just relying on camera poses or 3D points viewing angles (Section \ref{sec:camerasel}). 
\item An energy term to favor symmetric pairs and better compensate image noise while computing the gradient flow (Section \ref{sec:camerasel}).
\item An occlusion-aware mask term to explicitly identify, for each pixel, which part of the neighborhood has to be considered, during mesh refinement, to compute the similarity measure and its gradient (Section \ref{sec:occl}). 
\end{itemize}

%TODO aggiungere citazione a Fiorenti
\section{Related works}
Mesh refinement is a case of surface evolution.
Surface evolution methods are framed into the variational framework formalized by Hermosillo \etal \cite{hermosillo2002variational}.
Early works represent the model by level set, \ie, as the zero crossing of a function $f:\mathbf{R}^3\rightarrow\mathbf{R}$ \cite{faugeras2002variational,jin2002variational,yezzi2003stereoscopic,fuhrmann2014floating,solem2005geometric,yoon2010joint,pons2007multi}.
Faugeras and Keriven \etal \cite{faugeras2002variational} define the level set by means of a partial differential equation of $f$ such that a point on the surface moves proportionally to the photo-consistency of its neighborhood. Jin \etal \cite{jin2002variational} and Yoon \etal \cite{yoon2010joint} extend this approach to cope with specular reflection.
While these methods integrate the photometric measure in the 3D domain, Yezzi and Soatto \cite{yezzi2003stereoscopic} show that integrating this measure on the image domain yields to more accurate results.
Solem \etal \cite{solem2005geometric} and Pons \etal \cite{pons2005modelling,pons2007multi} replace the partial differential equation with a gradient leading to more robust mesh evolution.
Finally, Fuhrmann \etal \cite{fuhrmann2014floating} adapts the considered neighborhood around surface points according to their scale. 
Even if level set methods achieve accurate results, the evolution process is not always easy to track and understand.

Differently from level set methods, mesh refinement algorithms directly represent the surface as a 2D mesh embedded into a 3D space \cite{hiep2009towards,zaharescu2007transformesh,delaunoy_et_al_08,gargallo2007minimizing,delaunoy2011gradient,vu2011large}: given an initial rough mesh of the scene, they move the position of its vertices to obtain a more faithful model. 
Vu \etal \cite{hiep2009towards} discretized the continuous level set formulation of \cite{pons2007multi} to work directly with with meshes. 
Delaunoy \etal \cite{delaunoy2011gradient}  extended this method to take into account occlusions and in \cite{delaunoy2014photometric} they jointly optimize the surface and the camera in a bundle adjustment fashion.
Li \etal \cite{li2015detail} proposed an improved smoothness term of the energy function to output very smooth surfaces keeping the details of the scene.
Recently, Li \etal \cite{li2016efficient} simplify the mesh, decreasing the resolution where few vertices are sufficient to capture the structure of the scene, without affecting the accuracy significantly.
In \cite{romanoni2017mesh}, photometric refinement is coupled with a moving object detection method to avoid using their image areas to refine the static model of the scene.
Finally, two mesh refinement approaches \cite{blaha2017semantically,romanoni2017multi} exploit the semantic 2D segmentation of the images to improve the robustness of the refinement process, especially where two objects of different classes are adjacent.

\paragraph{Mesh Refinement}
\label{sec:ref}
Mesh refinement takes as input an initial mesh which is a rough model of the scene and a set of images capturing the scene.
The most popular approach, proposed in \cite{vu_et_al_2012}, minimizes an energy function $E$:
\begin{equation}
\label{eq:en}
E = E_{\textrm{photo}} + E_{\textrm{smooth}},
\end{equation}
where  $E_{\textrm{photo}}$ represents  the photo-consistency error of the model with respect to the images, and $E_{\textrm{smooth}}$  enforces the smoothness of the surface.

\begin{figure}[t]
\centering
  \def\svgwidth{0.7\columnwidth}
  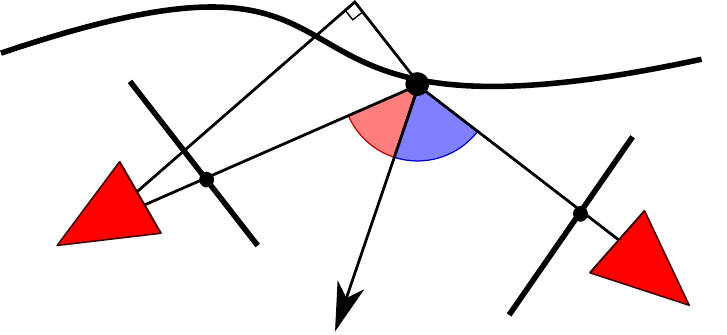
\caption{Variables involved in the photometric refinement process}
\label{fig:cameraproj}
\end{figure}

To minimize the term $E_{\textrm{photo}}$, the mesh refinement procedure applies gradient descent.
Let's consider two images $I$ and $J$, and a point $x$ belonging to the surface $\mathit{S}$ (Fig. \ref{fig:cameraproj}); we define the error function $err_{A, B}(x)$ that decreases if the similarity between the patch around the projection of $x$ in  $A$ and $B$ increases, e.g., in our case, the negative ZNCC of the $5x5$ pixels neighborhood.
The energy $E_{\textrm{photo}}$ in Equation \eqref{eq:en} is defined as:
\begin{equation}
\label{eq:energy_photo}
% \begin{split}
  E_{\textrm{photo}} = E(\mathit{S})  = \sum_{i,j}\int_{\Omega^{\textrm{S}}_{i,j}} err_{I, I_{ij}^{\mathit{S}}}(x_i)\textrm{d}x_i   = \sum_{i,j} \mathcal{E}^{im}_{ij}(x),
% \end{split}
\end{equation}
where  $I_{ij}^{\mathit{S}}$ represents the reprojection of the image from the $j$-th camera onto image $I$ through the surface $\mathit{S}$, and $\Omega^{\textrm{S}}_{i,j}$ is the image region where the reprojection is defined.
Now, let $X_i \in \mathbb{R}^3$ be a vertex of the surface mesh $\mathit{S}$, and $\phi_i(x)$ be the barycentric coordinates of a surface point $x$ in the triangle with vertex $X_i$.
The discrete gradient of $E_{\textrm{photo}} = E(\mathit{S})$ computed for a vertex $X_i$ is:
\begin{align}
\begin{split}
  \frac{\textrm{d}E(\mathit{S})}{\textrm{d}X_i}& =  \int_{\mathit{S}} \phi_i(x) \nabla E(S) \textrm{d}x = \int_{\mathit{S}} \phi_i(x) \nabla (\sum_{i,j} \mathcal{E}^{im}_{ij}(x)) \textrm{d}x =\\
  & = \int_{\mathit{S}} \phi_i(x) \sum_{i,j} \nabla \mathcal{E}^{im}_{ij}(x)\textrm{d}x.
\end{split}
\end{align}
By changing the variable of integration form $x$ to $x_i$ it is possible to integrate the energy over the image $I$ instead of over the surface $\mathit{S}$.
Let define $\overrightarrow{n}$ as the normal at $x$ pointing outward the surface $\mathit{S}$, $x_i$ the projection of x into the $I$ image, $\mathbf{d}_i$ as the vector from camera $i$ to $x$, $z_i$ as the depth of $x$ in camera $i$ (see Fig. \ref{fig:cameraproj}); with the change of variable $\textrm{d}x_i = -\overrightarrow{n}^T \mathbf{d}_i \textrm{d}x/z_i^3$ \cite{pons2007multi} we obtain:
\begin{equation}
\label{eq:final}
  \frac{\textrm{d}E(\mathit{S})}{\textrm{d}X_i} = 
  \sum_{i,j} \int_{\Omega^{\textrm{S}}_{i,j}} 
  \phi_i(x)  \nabla \mathcal{E}^{im}_{ij}(x)\frac{z_i^3}{\overrightarrow{n}^T \mathbf{d}_i }\overrightarrow{n} \textrm{d}x_i.
\end{equation}
% where $\Omega_{i,j}$ represents the surface region that induces the projection of image $J$ into the image $I$.

 To define which pairs $(i,j)$ are adopted to compute the gradients, the most widespread methods \cite{tola2012efficient,vu_et_al_2012,VuPhD011} leverage on 3D points correspondences estimated by the Structure from Motion method adopted to calibrate the cameras. For each camera $i$ it chooses the camera $j$ with the highest number of common 3D points with a reasonable parallax (\eg, in \cite{tola2012efficient} between $10\degree$ and $30\degree$).

Finally, the evolution process is complemented by the umbrella operator \cite{wardetzky2007discrete}, which moves each vertex in the mean position of its neighbors; this approximates the Laplace-Beltrami operator, and it minimizes the energy term $E_{\textrm{smooth}}$.

\begin{figure}[t]
\begin{center}
\centering
  \def\svgwidth{0.6\columnwidth}
  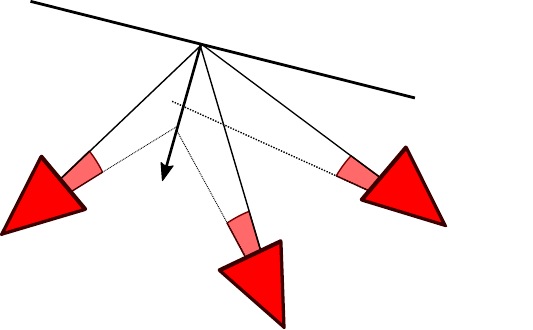
\caption{Example of symmetric ($C_{ref}$ and $C_1$) and non symmetric cameras ($C_{ref}$ and $C_2$) with respect to the surface normal $n$.}
\label{fig:sym}
\end{center}
\end{figure}

%======================================================================
%======================================================================
%======================================================================
\section{Model-based Camera Pairs Selection}
\label{sec:camerasel}
One of the most relevant aspect to effectively minimize $E_{photo}$ is the choice of the camera pairs $(i,j)$.
Instead of basing this choice on just 3D sparse points estimated by Structure from Motion, we propose to exploit the knowledge of the initial 3D mesh to find the camera pairs  having a good trade-off between reasonable parallax and image overlap. 

We first define a term $E_{p}^{i, j}$ to represent the quality of the parallax between camera $i$ and $j$.
Given the initial surface $\mathit{S}$, the camera centers, $c_i$ and $c_j$, and a point $x\in \mathit{S}$, let's define the parallax as the angle $\theta_{i,x,j} = \angle(c_{i}x, c_{j}x)$.
We compute the average parallax as:
\begin{equation}
 P_{i,j} = \frac{1}{A(\mathit{\Omega^{\textrm{S}}_{i,j}})}  \int_{\Omega^{\textrm{S}}_{i,j}} \theta_{i,\Pi_i^{-1}(x_i),j} \textrm{d}x_i
\end{equation}
where $A(\cdot)$ represents the area of an image region, in this case the region $\Omega^{\textrm{S}}_{i,j}$.
To define the reference parallax, we recall that small angles induce good overlap between patches, while larger angles induce more stability in the refinement process. 
Tola \etal  \cite{tola2012efficient} choose a small parallax between $10\degree$ and $30\degree$ to avoid erroneous image warping caused by occlusions.
In our case, however, we know the geometry, and we explicitly handle occlusions, therefore we prefer a larger reference angle. In \cite{morreale2018predicting} the parallax ranges around $40\degree$ to $70\degree$; in \cite{nocerino2014accuracy} and in \cite{wackrow2011minimising} the convergent angle, which is strictly related to the parallax, is chosen respectively as $50\degree$ and $45\degree$.  
Among these values, we experimentally choose the reference parallax to be $50\degree$. Therefore, we define:
\begin{equation}
E_{p}^{i, j} = - e^{\left(-\frac{(P_{i,j}-50\degree)^2 }{2 \cdot \sigma_p^2}\right)},
\end{equation}
where we put the variance $\sigma_p = 45\degree$.

To favor camera pairs with a similar resolutions and thus inducing coherent refinement, we introduce the resolution term $E_{R}^{i, j}$.
Let $\rho_i = \frac{||c_{i}x_i||}{f_i}$ and $\rho_j =\frac{||c_{j}x_i||}{f_j}$ be the distances of point $x_j$ from the two cameras respectively, normalized with respect to the corresponding focal length $f_i$ and $f_j$.
We define:
\begin{equation}
 \rho_{(i,j))} = \frac{\left| \rho_i - \rho_j\right|}{||c_{i}x_i||},
\end{equation}
as the normalized discrepancy of the resolutions of the two images with respect to the length of $c_{i}x_i$ ray.
We compute the average of these values as:
\begin{equation}
 R_{i,j} = \frac{1}{A(\mathit{\Omega^{\textrm{S}}_{i,j}})}  \int_{\Omega^{\textrm{S}}_{i,j}} \rho_{(i,j))} \textrm{d}x_i.
\end{equation}

To favor similar resolutions we define:
\begin{equation}
E_{r}^{i, j} = - e^{\left(-\frac{(R_{i,j}-0)^2 }{2 \cdot \sigma_r^2}\right)},
\end{equation}
where we put the variance $\sigma_r = 0.25$, which represents a resolution discrepancy of $25\%$.

Finally, to take into account overlap, we define $E_{o}^{i, j}$ as:
\begin{equation}
E_{o}^{i, j} = - \frac{A(\mathit{\Omega_{i,j}})}{A(I_{i})}.
\end{equation}

\paragraph{Symmetry Term} In most cases,  these two terms provide a fair evaluation of the camera pair quality with respect to the mesh refinement problem.
However, in some cases, they are not sufficient to discriminate among different camera pairs. For instance, in Fig. \ref{fig:sym}, the surface $\mathit{S}$ is perceived by a reference camera $C_{ref}$ and by two other cameras $C_1$ and $C_2$. 
The surface is entirely visible by the three cameras, \ie, the overlap is 100\%, and the baselines $C_{ref}-C_1$ and $C_{ref}-C_2$ have very similar values; by relying on just $E_{p}$ and $E_{o}$,  both pairs $C_{ref}, C_1$ and  $C_{ref}, C_2$ are considered equally good (or equally bad). 

To overcome this issue, we evaluate a third term that favors camera pairs with points of view symmetric with respect to the scene. 
Intuitively, since the surface $\mathit{S}$ evolves along its normal $\overrightarrow{n}$ and assuming the images affected by Gaussian noise on the image plane, if we compute the gradient between $I_{ref}$ captured by $C_{ref}$ and $I_2$ captured by $C_2$, similar noise in $I_{ref}$ and in $I_{2}$ (inducing an uncertainty angle $\tau$) translate into significantly different gradient noises along $\overrightarrow{n}$. Instead, if we consider $I_{ref}$ and $I_1$ the same noise affects similarly the gradients along $\overrightarrow{n}$. Statistically, in the former case, the noise accumulates as the gradients are computed, while, in the latter, they likely compensate each other.

In addition to parallax and overlap  we then evaluate a symmetry term $E_{s}^{i, j}$ with respect to the surface normal. To do so we define the oriented angle difference (OAD):
% \begin{equation}
$
{\delta_{i,x,j} := sign \cdot \frac{1}{2}\left[\angle\left(c_{i}x,\overrightarrow{n}(x)\right) - \angle\left(c_{j}x,\overrightarrow{n}(x)\right)\right] ,} 
$
% \end{equation}
where $\overrightarrow{n}(x)$ is the normal of the surface $\mathit{S}$ at $x$ and, if the $c_{i}x$ and $c_{j}x$
belong to the same half-space defined by the plane parallel to $\overrightarrow{n}$ through $x$, then $sign = 1$,  otherwise $sign = -1$.
The OAD average on the surface is computed as:
\begin{equation}
 S_{i,j} = \frac{1}{A(\mathit{\Omega^{\textrm{S}}_{i,j}})}  \int_{\Omega^{\textrm{S}}_{i,j}} \delta_{i,\Pi_i^{-1}(x_i),j} \textrm{d}x_i,
\end{equation}
and, the novel energy term $E_{s}^{i, j}$ is computed as:
% \begin{equation}
${
E_{s}^{i, j} = - e^{\left(-\frac{(S_{i,j}-0\degree)^2}{2 \cdot \sigma_s^2}\right)},}$
% \end{equation}
where we put experimentally the variance $\sigma_s = \sigma_p = 45\degree$.
This  term is combined with $E_{p}^{i, j}$ and $E_{o}^{i, j} $ to define the energy function $E_{\text{BPV}}(i,j)$ for a pair of cameras $C_i$ and $C_j$:
\begin{equation}
E_{\text{BPV}}^{i, j} = \mu_1 \cdot E_{p}^{i, j} +  \mu_2 \cdot E_{o}^{i, j} +  \mu_3 \cdot E_{s}^{i, j} +  \mu_4 \cdot E_{r}^{i, j},
\end{equation}
where $\mu_1$, $\mu_2$ and $\mu_3$ are three  coefficient weighting the  contribution of the three energy term (in our case $\mu_1=0.25$, $\mu_2=0.25$, $\mu_3=0.5$ and $\mu_4=0.25$).

%  \begin{algorithm}[t]
%  \caption{Initial Camera Pairs Selection}
%  \label{alg:init}
%  \begin{algorithmic}[1]
%  \renewcommand{\algorithmicrequire}{\textbf{Input:}}
%  \renewcommand{\algorithmicensure}{\textbf{Output:}}
%  \REQUIRE 	\hfill \\
%  			$\mathbb{C}$, set of cameras, 
%  			$\mathcal{M}$ 3D mesh
%  \ENSURE  \hfill \\
%  			$\mathbb{P}^0$, selected camera pairs
%   \FOR {$i \in \mathbb{C}$}
%   \STATE $\mathbb{P}^0 \leftarrow (i, \argmin_j\left\{E_{\text{BPV}}^{i, j}\right\})$
%   \ENDFOR
%  \RETURN $\mathbb{P}^0$
%  \end{algorithmic}
%  \end{algorithm}

 \begin{algorithm}[t]
 \caption{Camera Pairs Selection}
 \label{alg:select}
 \begin{algorithmic}[1]
 \renewcommand{\algorithmicrequire}{\textbf{Input:}}
 \renewcommand{\algorithmicensure}{\textbf{Output:}}
 \REQUIRE \hfill \\
 			$\mathbb{C}$, set of $N_{cam}$ cameras,
 			$\mathcal{M}$ 3D mesh, 
 			$\mathbb{P}^0$            
 \ENSURE  \hfill \\
 			$\mathbb{P}$, set of camera pairs\\
  \textbf{Algorithm:}\\
  Initial Camera Pairs Selection:
  \FOR {$i \in \mathbb{C}$}
  \STATE $\mathbb{P}^0 \leftarrow (i, \argmin_j\left\{E_{\text{BPV}}^{i, j}\right\})$
  \ENDFOR
  Camera Pairs Selection:
  \STATE $changed = true$
  \STATE $camUsed = \left\{{\emptyset}_1,\dots, {\emptyset}_{N_{cam}} \right\}$
  \STATE $\mathbb{P}^{\textrm{Prev}} = \mathbb{P}^0$
  \STATE $E_{\text{ref}} = \sum_{(i,j)\in {\mathbb{P}}^0}E_{\text{BPV}}^{i, j}$
  \STATE $E_{\text{new}} = E_{\text{ref}}$
  \WHILE {$E_{\text{new}} > 0.9 \cdot E_{\text{ref}}$ \AND $changed$}
  	\STATE $changed = false$ 
    \STATE $\mu_{prev} = \mu^{\mathbb{P}^0} $ , $\sigma_{prev} = \sigma^{\mathbb{P}^0} $
    \FOR {$i \in \mathbb{C}$}
    	\STATE $\bar{\mathbb{P}}^{\textrm{Cur}} = \mathbb{P}^{\textrm{Prev}}$  
        \STATE remove $(i,j_{\text{old}})$ from $\bar{\mathbb{P}}^{\textrm{Cur}}$
        \STATE $j_{new} = \argmin_j\left\{E_{\text{BPV}}^{i, j} s.t. (i, j) \notin camUsed_i\right\}$
        \STATE $\bar{\mathbb{P}}^{\textrm{Cur}}\leftarrow (i,j_{new})$
        \STATE $\mu_{cur} = \mu^{\textrm{Cur}}$
        \STATE $\sigma_{cur} = \sigma^{\textrm{Cur}}$
      \IF {$\mu_{cur} > \mu_{prev}$ \AND $\sigma_{cur} < \sigma_{prev}$}
        \STATE ${\mathbb{P}}^{\textrm{Cur}} = \bar{\mathbb{P}}^{\textrm{Cur}}$
        \STATE $camUsed_i = camUsed_i \cup j_{\text{new}}1$;
  		\STATE $E_{\text{new}} = E_{\text{new}} + E_{\text{BPV}}^{i, j_{new}} - E_{\text{BPV}}^{i, j_{old}}$
        \STATE $changed = true$
        \STATE $\mu_{prev} = \mu_{cur}$
        \STATE $\sigma_{prev} = \sigma_{cur}$
      \ELSE
        \STATE ${\mathbb{P}}^{\textrm{Cur}} = \bar{\mathbb{P}}^{Prev}$
      \ENDIF
    \ENDFOR
  \ENDWHILE
 \RETURN $\mathbb{P}^{\textrm{Cur}}$
 \end{algorithmic}
 \end{algorithm}

\paragraph{Model Coverage}

A second relevant aspect when choosing a camera pair is model coverage. 
While the overlap term is related to the overlap among the images in the pair, in principle no terms discussed previously avoids all the camera pairs perceive, and therefore refine just a small portion of the mesh.
For this reason, we enforce  camera pair configurations providing good coverage.

We first initialize the camera pairs as follows.
Let $\mathbb{C}$ be the set of cameras and  $\mathbb{P}$  the set of camera pairs adopted for the refinement.
Our algorithm initializes the initial set $\mathbb{P}^0$ of best camera pairs computed leveraging on the previous energy as $\mathbb{P}^0 = \left\{(i, j) \, \forall i\in \mathbb{C} \, \textrm{s.t.} \, j = \argmin_j\left\{E_{\text{BPV}}^{i, j}\right\} \right\}$, \ie, with the best pair for each camera.

To enforce model coverage, the idea is to perturb this initial set of pairs $\mathbb{P}^0$.
In the first step we define the model coverage as the average number of camera pairs in which all the facets are visible.
Let $\mathbb{F}$ be the set of facets and let define a visibility function $v_f^{i,j}$ of facet $f\in \mathbb{F}$ with respect to cameras $i$ and $j$, \ie $v_f^{i,j} = 1$ if is visible from both cameras, $v_f^{i,j} = 0$ otherwise.
Then, the global visibility of $f$ is  $V_f = \sum_{i,j}v_f^{i,j}$ and the coverage of the whole mesh is represented by $C_{\mathbb{P}} = \left\{ V_f, \forall f \in F\right\}$.

The second step aims at replacing camera pair $(i,j)$ with a reasonable pair $(i,k)$ that, even at the cost of a small decrease of energy $E_{\text{BPV}}$ it improves the model coverage.

To do so, given the initial set $\mathbb{P}^0$, we compute
% \begin{equation}
${\mu^{\mathbb{P}^0} = \mathbb{E}(C_{\mathbb{P}^0})}$  and
${\sigma^{\mathbb{P}^0} = {stddev}(C_{\mathbb{P}^0})}$
% \end{equation}
For each camera $i$ we compare the current camera pair $(i,j_1)\in\mathbb{P}^0$ with the second best camera pair $(i,j_2)$ among the pairs $(i,\cdot)$. 
If $(i,j_2)$ increases the mean coverage $\mu^{\mathbb{P}^0}$ and decreases $\sigma^{\mathbb{P}^0}$ we try to switch $(i,j_1)$ and $(i,j_2)$ so to obtain a new set $\bar{\mathbb{P}}^1$. 
If the sum of the energies $\sum_{(i,j)\in \bar{\mathbb{P}}^1}E_{\text{BPV}}^{i, j} < 0.9 \cdot \sum_{(i,j)\in {\mathbb{P}}^0}E_{\text{BPV}}^{i, j}$ then the pair change is successful, \ie, $\mathbb{P}^1 = \bar{\mathbb{P}}^1$, otherwise $\mathbb{P}^1 = {\mathbb{P}}^0$
We iterate this process until no further change happens (Algorithm \ref{alg:select}).

Let notice that in the previous process we considered ony one camera $j$ for each reference camera $i$. This does not represent a limitation of the algorithm but a choice to have a fair comparison with the baselines of Tola \etal \cite{tola2012efficient} and  our implementation of Vu \etal \cite{vu_et_al_2012} that compare one camera for each reference too.
We refer to the experimental section for a more detailed discussion.

\begin{figure}[t]
\centering
\setlength{\tabcolsep}{1px}
\begin{tabular}{ccc}
\includegraphics[width=0.3\columnwidth]{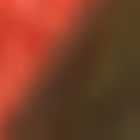}&
\includegraphics[width=0.3\columnwidth]{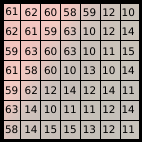}&
\includegraphics[width=0.3\columnwidth]{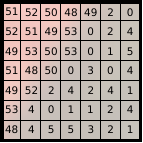}\\
(a) & (b) & (c)\\
\end{tabular}
\caption{Patch (a), Depth Patch (b) and Difference of depths patch (c)}
\label{fig:occlmPatch}
\end{figure}

\section{Occlusion Aware Masking}
\label{sec:occl}
For each pixel $p(x,y)$, the mesh refinement presented in Section \ref{sec:ref} aggregates the gradients of the similarity measure from the neighboring pixels in a squared patch $P$ (in our case with size $5$x$5$px) (Fig. \ref{fig:occlmPatch}(a)).

In most cases, all the pixels in $P$ contain information useful to refine the position of the 3D point corresponding to $p(x,y)$ by gradient descent.
However, in the case of occlusions, the squared patch $P$ contains information from regions of the scene not related to $p(x,y)$, which translates into errors in the gradient computation.
To tackle this issue, we rely on the depth map of the current 3D model of the scene. The idea is to find which pixels of $P$ have a depth similar and coherent to the pixel $p(x,y)$, and to use only them during the similarity gradient computation. In the following these pixels are named \emph{valid pixels}.

For each pixel $p(h,k)\in P$, we compute the depth values $\delta(h,k)$ as the camera to model distance (\ref{fig:occlmPatch}(b)) and its difference with respect to the depth of pixel $p(x,y)$ as $dd(h,k)= \delta(h,k) - \delta(x,y)$ (\ref{fig:occlmPatch}(c)).

Since abrupt depth discontinuities would induce very high variances on the whole patch $P$, in principle, the standard deviation of the $dd(h,k)$ values are not enough informative to classify the \emph{valid pixels}.

For this reason we propose the following procedure-
First, we cluster the pixels between those with a depth closer and those with a depth are farther to the depth of $p(x,y)$. In other words, we cluster all the values $dd(h,k)$ in two sets $DD_{min}$ and $DD_{max}$, which contain respectively the $dd(h,k)$ values closest to $0$ and closest to $max\{dd(h,k)\}$.

Then, we approximate a robust estimator $\hat{\sigma}$ of the standard deviation of the depths of the valid pixels.
Assuming the pixels in $DD_{min}$ to be likely \emph{valid}, and the highest depth variances usually corresponding to the outer part of the patch, we compute $\hat{\sigma}$ as the value $dd(h,k)\in DD_{min}$ of the (spatially) farthest pixel w.r.t. the patch center $(x,y)$.
A pixel $p(h,k)$ is valid when $|dd(h,k) - \hat{\sigma}| < 10|dd(h,k) - max\{dd(h,k)\}|$

\begin{figure}[t]
\centering
\setlength{\tabcolsep}{1px}
\begin{tabular}{cc}
\includegraphics[width=0.2\textwidth]{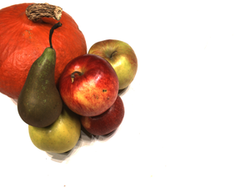}&
\includegraphics[width=0.2\textwidth]{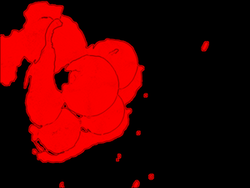}\\
\end{tabular}
\caption{Number of pixels considered for mesh refinement after occlusion masking }
\label{fig:occlm}
\end{figure}

In Fig.  \ref{fig:occlm} we illustrate the first iteration of mesh refinement for DTU sequence 63: the intensity of red represents the number of pixels considered in the gradient computation. It is possible to notice that the number of pixels considered decreases approaching to the mesh discontinuities.

\begin{figure}[tp]
\centering
\setlength{\tabcolsep}{1px}
\begin{tabular}{cc}
\includegraphics[width=0.4\columnwidth]{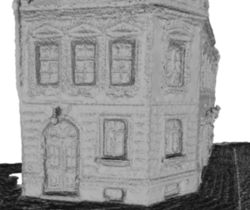}&
\includegraphics[width=0.4\columnwidth]{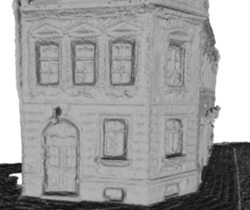}\\
(a) without mask refinement & (b) with mask refinement\\
\end{tabular}
\caption{The effect of the masking refinement}
\label{fig:resDtu6}
\end{figure}

\begin{figure}[tp]
\centering
\setlength{\tabcolsep}{1px}
\begin{tabular}{cc}
\includegraphics[width=0.4\columnwidth]{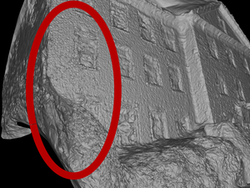}&
\includegraphics[width=0.4\columnwidth]{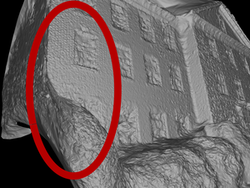}\\
(a) Vu \etal~\cite{vu_et_al_2012}&
(b) Proposed\\
\end{tabular}
\caption{The effect of the coverage algorithm}
\label{fig:resSouth}
\end{figure}

\begin{table}[!ht]
    \centering
    \scriptsize
 \setlength{\tabcolsep}{4px}
\renewcommand{\arraystretch}{1.02} % Default value: 1
    \caption{Accuracy (model-to-GT distance) and Completeness  (GT-to-model distance) of the refined model w.r.t. the state-of-the-art on the DTU dataset. }
    \label{tab:DTU}
    \begin{tabular}{lllccccc}
    &&& \cite{campbell2008using} &  \cite{fu10} &  \cite{tola2012efficient}    &   \cite{vu_et_al_2012} &Proposed  \\
    \hline
    \multirow{ 4}{*}{4}&\multirow{ 2}{*}{Acc.} & Mean&0.7265&0.7947&0.3202&0.3254&\textbf{0.3171}\\
    & & Median &0.4014&0.3016&\textbf{0.1888}&0.2028&0.1972\\
    & \multirow{ 2}{*}{Compl.} & Mean&0.6054&\textbf{0.6014}&0.7791&0.7988&0.7936\\
    & & Median &0.4257&0.3644&\textbf{0.3245}&0.3463&0.3370\\
    \hline
    \multirow{ 4}{*}{6}&\multirow{ 2}{*}{Acc.} & Mean&1.0727&0.6660&0.3681&0.3666&\textbf{0.3433}\\
    & & Median &0.4568&0.2628&\textbf{0.1993}&0.2095&\textbf{0.1990}\\
    & \multirow{ 2}{*}{Compl.} & Mean&\textbf{0.4281}&0.4739&0.5208&0.5221&0.5095\\
    & & Median &0.3454&0.3568&0.3335&0.3427&\textbf{0.3273}\\
    \hline
    \multirow{ 4}{*}{15}&\multirow{ 2}{*}{Acc.} & Mean&0.9829&0.8544&0.4771&0.4614&\textbf{0.4436}\\
    & & Median &0.5588&0.4379&0.2866&0.2754&\textbf{0.2689}\\
    & \multirow{ 2}{*}{Compl.} & Mean&0.3243&0.5028&0.6906&\textbf{0.6709}&0.6737\\
    & & Median &\textbf{0.2606}&0.3876&0.4203&0.3973&0.3981\\
    \hline
    \multirow{ 4}{*}{18}&\multirow{ 2}{*}{Acc.} & Mean&1.3916&0.9665&0.5125&\textbf{0.4918}&\textbf{0.4696}\\
    & & Median &0.6793&0.4247&0.2603&0.2592&\textbf{0.2547}\\
    & \multirow{ 2}{*}{Compl.} & Mean&\textbf{0.3640}&0.4922&0.8996&0.8763&0.8742\\
    & & Median &\textbf{0.2899}&0.3715&0.3937&0.3740&0.3720\\
    \hline
    \multirow{ 4}{*}{24}&\multirow{ 2}{*}{Acc.} & Mean&3.4509&0.7518&0.3941&0.3871&\textbf{0.3802}\\
    & & Median &0.5255&0.3141&0.2659&0.2651&\textbf{0.2632}\\
    & \multirow{ 2}{*}{Compl.} & Mean&\textbf{0.4010}&0.4827&0.8512&0.8293&0.8300\\
    & & Median &\textbf{0.2954}&0.3691&0.4339&0.4119&0.4124\\
    \hline
    \multirow{ 4}{*}{36}&\multirow{ 2}{*}{Acc.} & Mean&0.5972&0.6270&0.3125&0.2859&\textbf{0.2801}\\
    & & Median &0.2317&0.2778&0.2007&\textbf{0.1831}&\textbf{0.1831}\\
    & \multirow{ 2}{*}{Compl.} & Mean&\textbf{0.4622}&0.6101&1.0331&1.0093&1.0070\\
    & & Median &\textbf{0.2317}&0.2930&0.2856&0.2527&0.2533\\
    \hline
    \multirow{ 4}{*}{63}&\multirow{ 2}{*}{Acc.} & Mean&2.4241&2.3992&0.9082&0.8461&\textbf{0.7836}\\
    & & Median &0.2782&1.1192&0.2711&0.2495&\textbf{0.2303}\\
    & \multirow{ 2}{*}{Compl.} & Mean&\textbf{0.4730}&0.6401&0.7189&0.7159&0.7158\\
    & & Median &\textbf{0.2782}&0.3849&0.2985&0.2916&0.2924\\
    \hline
    \multirow{ 4}{*}{106}&\multirow{ 2}{*}{Acc.} & Mean&0.5918&0.7881&0.3028&0.2844&\textbf{0.2765}\\
    & & Median &0.2793&0.3028&0.1902&0.1846&\textbf{0.1821}\\
    & \multirow{ 2}{*}{Compl.} & Mean&\textbf{0.6902}&0.7004&0.9950&0.9936&0.9935\\
    & & Median &\textbf{0.2793}&0.3244&0.3256&0.3220&0.3226\\
    \hline
    \multirow{ 4}{*}{110}&\multirow{ 2}{*}{Acc.} & Mean&3.4509&1.0922&0.7378&0.6867&\textbf{0.6674}\\
    & & Median &0.5255&0.3802&0.2314&0.2237&\textbf{0.2222}\\
    & \multirow{ 2}{*}{Compl.} & Mean&\textbf{0.4010}&0.5547&0.5675&0.5872&0.5892\\
    & & Median &\textbf{0.2954}&0.4041&0.4134&0.4238&0.4272\\
    \hline
    \multirow{ 4}{*}{114}&\multirow{ 2}{*}{Acc.} & Mean&0.6104&0.5789&0.2734&\textbf{0.2696}&0.2714\\
    & & Median &0.29700&0.2616&0.1835&\textbf{0.1789}&0.1802\\
    & \multirow{ 2}{*}{Compl.} & Mean&\textbf{0.3544}&0.4001&0.3895&0.3872&0.3878\\
    & & Median &\textbf{0.2948}&0.3191&0.2999&0.2964&0.2970\\
    \hline
    \multirow{ 4}{*}{118}&\multirow{ 2}{*}{Acc.} & Mean&5.5335&0.65.25&0.3093&0.2982&\textbf{0.2911}\\
    & & Median &4.0794&0.2922&0.1946&0.1913&\textbf{0.1897}\\
    & \multirow{ 2}{*}{Compl.} & Mean&\textbf{0.3682}&0.5489&0.7389&0.7416&0.7399\\
    & & Median &\textbf{0.2856}&0.3296&0.3301&0.3302&0.3290\\
    \hline
    \multirow{ 4}{*}{122}&\multirow{ 2}{*}{Acc.} & Mean&0.6367&0.6621&0.3049&0.2920&\textbf{0.2884}\\
    & & Median &0.3276&0.2967&0.1978&0.1920&\textbf{0.1912}\\
    & \multirow{ 2}{*}{Compl.} & Mean&\textbf{0.3682}&0.4576&0.6435&0.6441&0.6400\\
    & & Median &\textbf{0.2856}&0.3281&0.3278&0.3256&0.3226\\
    \end{tabular}
\end{table}

\begin{table*}[tbh]
    \centering
 \setlength{\tabcolsep}{5px}
    \caption{Ablation study for the 12 sequences of the DTU dataset. Quantitative comparison according to the overall Average Ranking of the 3D model errors on the 12 sequences and the overall average of accuracy and completeness errors (Average Error Value)}
    \label{tab:DTU_ablation}
    \begin{tabular}{lccccccccc}%PSO
    &  PC & SC & OC & SPC  & OPC & OSC & OSP & OSPC   & Proposed\\
    \hline
    {Average Position}& 6.2 & 5.2 & 4.9 & 4.9 & 6.1 & 4.8 & 5.8 &  \textbf{3.3} & \textbf{3.3}\\
    {Average Error Value}& 0.4295 & 0.4258 & 0.4285 & 0.4276 & 0.4258 & 0.4281 & 0.4280 & {0.4223} &\textbf{0.4212}\\
    \end{tabular}
\end{table*}

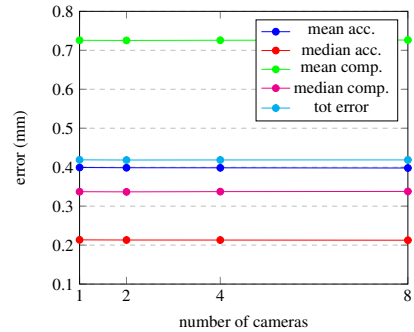
\begin{figure}[tp!h]
 
\centering
\resizebox {0.3\textwidth} {!} {
  \begin{tikzpicture}
  \begin{axis}[
      enlargelimits=false,
      ylabel={error (mm)},
      xlabel={number of cameras },
       xmin=1, xmax=8,
       ymin=0.1, ymax=0.8,
      xtick={1,2,4,8},
      ytick={0,0.1,0.2,0.3,0.4,0.5,0.6,0.7,0.8},
      legend pos=north east,
      ymajorgrids=true,
      grid style=dashed,
  ]
  \addplot[
      color=blue,
      mark=oplus*]
  table[x index=0,y index=1,col sep=comma]
  {data/multi-cam.txt};
  \addplot[
      color=red,
      mark=oplus*]
  table[x index=0,y index=2,col sep=comma]
  {data/multi-cam.txt};
  \addplot[
      color=green,
      mark=oplus*]
  table[x index=0,y index=3,col sep=comma]
  {data/multi-cam.txt};
  \addplot[
      color=magenta,
      mark=oplus*]
  table[x index=0,y index=4,col sep=comma]
  {data/multi-cam.txt};
  \addplot[
      color=cyan,
      mark=oplus*]
  table[x index=0,y index=5,col sep=comma]
  {data/multi-cam.txt};
  \legend{mean acc.,median acc.,mean comp., median comp.,tot error}
  \end{axis}
  \end{tikzpicture}
}

\caption{Errors for different number of camera compared to the reference image}
\label{fig:multicam}
\end{figure}

\begin{table}[t]
    \centering
\renewcommand{\arraystretch}{1.00} % Default value: 1
    \caption{Results on the fountain-P11 dataset (depth errors in $m$).}
    \label{tab:fount}
    \begin{tabular}{lccc}
    &{Initial Mesh}  &  {\cite{vu_et_al_2012}}&{Proposed } \\
    \hline
    MAE &0.001513 &  0.001360 & \textbf{0.001199} \\
    RMSE & 0.03890 &0.03688 &\textbf{0.03462} \\
    \end{tabular}
\end{table}

\section{Experiments}
We tested our refinement method against 12 sequences of the DTU dataset \cite{jensen2014large}, the fountain-P11 \cite{strecha2008} and the Southbuilding \cite{hane2013joint} sequences.
In Table \ref{tab:DTU} we reported the quantitative results on the DTU dataset against state of the are MVS methods.
As mentioned in the Section \ref{sec:camerasel} to have a fair comparison against Tola \etal and Vu \etal we choose one image for each reference camera. Indeed, both in Tola \etal and in our implementation of Vu \etal for each reference camera the second camera is chosen by relying on the knowledge of the visibility of the Structure from Motion 3D points. For the implementation of Vu \etal we choose the camera with the highest number of common points with a parallax between 20$\degree$ and 60$\degree$, while Tola \etal with a parallax between 10$\degree$ and 30$\degree$ and at least 0.8 scale difference between the corresponding DAISY descriptors.
In Figure \ref{fig:multicam} we tested our method comparing 1, 2, 4 or 8 cameras with respect to the reference image and the improvement obtained is almost negligible.

\begin{figure*}[tb]
\centering
\setlength{\tabcolsep}{1px}
\begin{tabular}{ccccc}
\multicolumn{5}{c}{DTU 18}\\
\includegraphics[width=0.185\textwidth,height=0.121\textwidth]{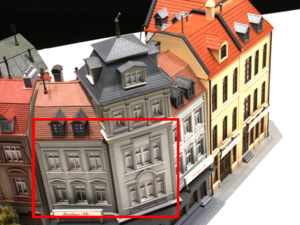}&
\includegraphics[width=0.185\textwidth,height=0.121\textwidth]{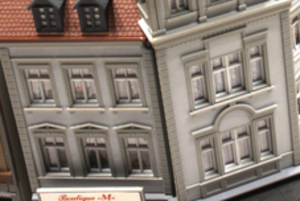}&
\includegraphics[width=0.185\textwidth,height=0.121\textwidth]{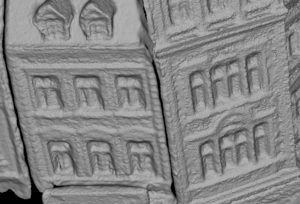}&
\includegraphics[width=0.185\textwidth,height=0.121\textwidth]{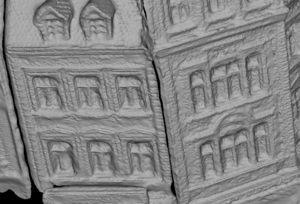}&
\includegraphics[width=0.185\textwidth,height=0.121\textwidth]{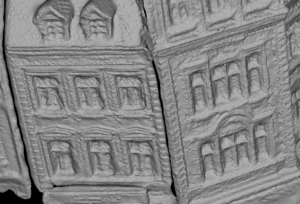}\\
\multicolumn{5}{c}{DTU 24}\\
\includegraphics[width=0.185\textwidth,height=0.121\textwidth]{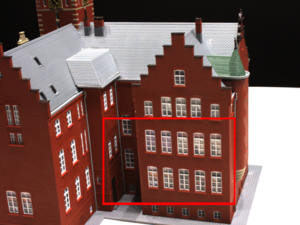}&
\includegraphics[width=0.185\textwidth,height=0.121\textwidth]{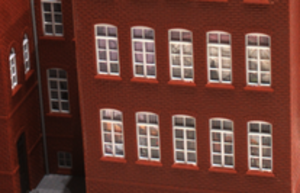}&
\includegraphics[width=0.185\textwidth,height=0.121\textwidth]{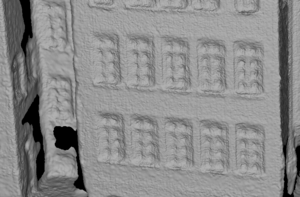}&
\includegraphics[width=0.185\textwidth,height=0.121\textwidth]{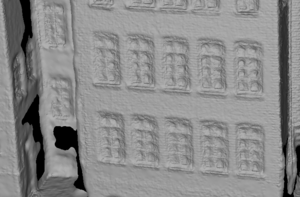}&
\includegraphics[width=0.185\textwidth,height=0.121\textwidth]{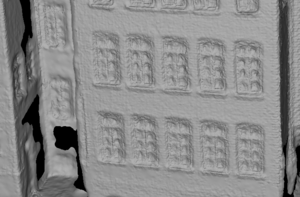}\\
\multicolumn{5}{c}{DTU 36}\\
\includegraphics[width=0.185\textwidth,height=0.121\textwidth]{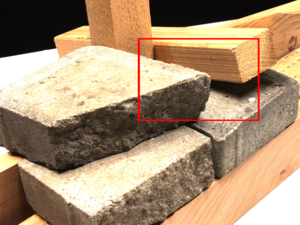}&
\includegraphics[width=0.185\textwidth,height=0.121\textwidth]{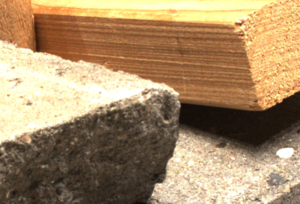}&
\includegraphics[width=0.185\textwidth,height=0.121\textwidth]{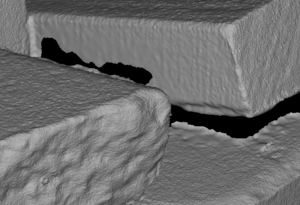}&
\includegraphics[width=0.185\textwidth,height=0.121\textwidth]{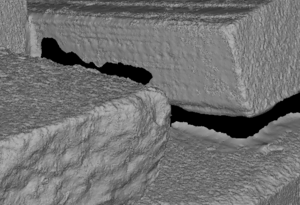}&
\includegraphics[width=0.185\textwidth,height=0.121\textwidth]{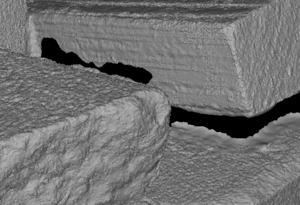}\\
\multicolumn{5}{c}{fountain-p11}\\
\includegraphics[width=0.185\textwidth,height=0.121\textwidth]{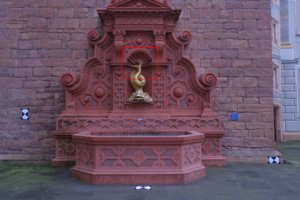}&
\includegraphics[width=0.185\textwidth,height=0.121\textwidth]{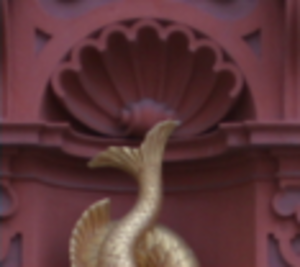}&
\includegraphics[width=0.185\textwidth,height=0.121\textwidth]{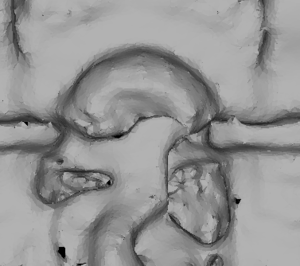}&
\includegraphics[width=0.185\textwidth,height=0.121\textwidth]{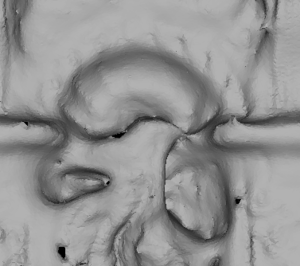}&
\includegraphics[width=0.185\textwidth,height=0.121\textwidth]{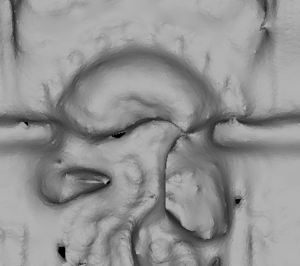}\\
\multicolumn{5}{c}{Southbuilding}\\
\includegraphics[width=0.185\textwidth,height=0.121\textwidth]{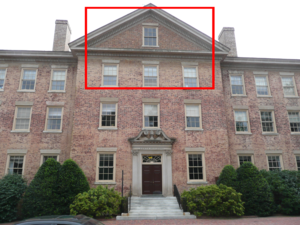}&
\includegraphics[width=0.185\textwidth,height=0.121\textwidth]{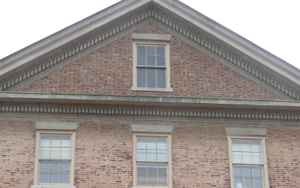}&
\includegraphics[width=0.185\textwidth,height=0.121\textwidth]{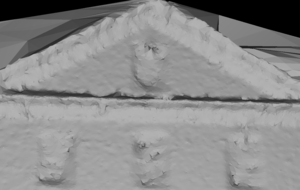}&
\includegraphics[width=0.185\textwidth,height=0.121\textwidth]{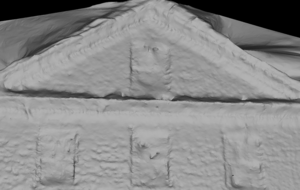}&
\includegraphics[width=0.185\textwidth,height=0.121\textwidth]{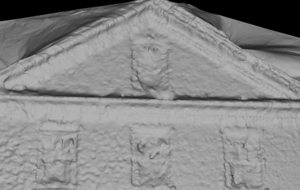}\\
% \multicolumn{4}{c}{castle-p30}\\
% \includegraphics[width=0.23\textwidth]{./images/castle}&
% \includegraphics[width=0.23\textwidth]{./images/castlep11or_000008}&
% \includegraphics[width=0.23\textwidth]{./images/castlep11sfm_000008}&
% \includegraphics[width=0.23\textwidth]{./images/castlep11model_000008}\\
RGB image & RGB detail & \cite{tola2012efficient}& \cite{vu_et_al_2012}&Proposed\\
\end{tabular}
\caption{Examples of refinement results with respect to the initial 3D mesh and the baseline refinement method }
\label{fig:res}
\end{figure*}

According to the procedure described in \cite{jensen2014large} we compared the distance (in mm) from refined 3D mesh to the ground truth point cloud  to compute the accuracy of the model and vice-versa to evaluate its completeness. In the table we list the mean and median values of such distances. In these experiments we used as initial mesh those extracted by the method of Tola \etal + Poisson Reconstruction.  Our method improves accuracy and completeness with respect to the baseline refinement \cite{vu_et_al_2012}.  
In Table \ref{tab:DTU_ablation} we show the ablation study on the DTU dataset and in Fig. \ref{fig:res} we reported some examples of the outcome of our mesh refinement. From the actual proposal (last column of Table \ref{tab:DTU_ablation}) we tested the refinement by using just subsets of the energy terms (P = Parallax, S = Symmetry, O = Overlap), with or without the coverage algorithm (represented by C) and with or without the masked refinement (represented by M).
We do not mention the resolution prior since, in our scenario, it does not affect the result. This outcome is expected and coherent because the cameras have approximately the same distant from the model and the same image resolution.
To have a better understanding of the performance of each version, we reported the average rank and the average error value. 
To compute the former, we ranked the algorithms for each row, \ie, for each error measure and each sequence, and we reported the average of these rankings.
The latter was also used in \cite{poms2018learning} and it averages all the accuracy and completeness mean and median values for each column.
In both cases, it is possible to notice that all the terms contribute to achieving a better outcome which is a good trade-off between accuracy and completeness.
Fig. \ref{fig:resDtu6} shows how the masking refinement produces smoother results, by limiting the effect of occlusions.

Table \ref{tab:fount} shows a further quantitative comparison evaluating the accuracy for the fountain-P11 dataset. We computed the Mean Absolute Error (MAE) and the Root Mean Squared Error (RMSE) as proposed in \cite{strecha2008}, \ie, by comparing the depth maps rendered from the reconstructed model and the GT mesh. Our refinement method improves the accuracy of the classical refinement method in \cite{vu_et_al_2012}.

The Southbuilding dataset does not provide a reference ground truth. However from  Fig. \ref{fig:res} and Fig. \ref{fig:resSouth} (and the images reported in the supplementary material with higher resolution) it is possible to notice that our refinement method produces an output with more details than\cite{vu_et_al_2012}, thanks to the effective choice of camera pairs, together with the masked refinement.
In particular, Fig. \ref{fig:resSouth} shows that the coverage procedure avoids refinement to focus only on one part of the model producing unbalanced mesh evolution (Fig. \ref{fig:resSouth}(a)).

It is worth noticing that the method we use to define the pairs of cameras and to mask the occlusions during the photometric refinement can be easily applied to any mesh refinement method without any relevant modification in the code.

\section{Conclusions and future works}
In this paper, we addressed two relevant issues of mesh refinement: the choice of camera pairs used to compute similarity gradients and the occlusion management.
We defined a model-based energy function to evaluate the overlap, the parallax the resolution and the symmetry among the camera pairs and we proposed a procedure to choose the set of pairs which provides a good trade-off between the defined energy and the model coverage.
We also proposed a novel strategy to mask the region of the patch adopted to compute the similarity measures and the gradients such that the influences of model occlusions ad discontinuities are  neglected or, at least, limited. 
As future work, we plan to propose a parallel version of the presented
refinement method that splits the mesh into several parts that can be processed independently. This allows exploring the level of parallelism depending on both the required mesh resolution and the computing platform. Moreover, we can optimize the energy-performance trade-off by leveraging the dynamic reconfiguration support offered in multi-core architectures~\cite{zoni_blackout2017,zoni_taco2019}.
We also intend to improve the camera selection exploiting the information recovered by the dense Multi-View Stereo method such as \cite{schonberger2016pixelwise}, which jointly estimate image depths and pixel-wise camera pairs.

{\small
\subsection*{\bfseries Acknowledgements}
This work was partially founded by EIT digital
}
\bibliographystyle{elsarticle-num}

\bibliography{biblioTotal}

\end{document}